\documentclass{article}

    \PassOptionsToPackage{numbers, compress}{natbib}

\usepackage[nonatbib,final]{neurips_2020}
\usepackage{xcolor}
\usepackage{graphicx}
\usepackage[utf8]{inputenc} 
\usepackage[T1]{fontenc}    
\usepackage[hidelinks]{hyperref}       
\usepackage{url}            
\usepackage{booktabs}       
\usepackage{amsfonts}       
\usepackage{amsmath}
\usepackage{nicefrac}       
\usepackage{microtype}      
\usepackage[textsize=tiny]{todonotes}
\usepackage{titlesec}
\titlespacing*{\section}{0pt}{0.7\baselineskip}{0.3\baselineskip}

\title{Towards Automated Satellite Conjunction Management with Bayesian Deep Learning}

\author{%
\begin{minipage}[t]{0.466\textwidth}
  \begin{center}
  Francesco Pinto\\
  \normalfont{University of Oxford}\\
  \texttt{francesco.pinto@eng.ox.ac.uk}
  \end{center}
\end{minipage}
  \And 
\begin{minipage}[t]{0.466\textwidth}
  \begin{center}
  Giacomo Acciarini\\
  \normalfont{University of Strathclyde}\\
  \texttt{giacomo.acciarini@gmail.com}
  \end{center}
\end{minipage}
  \And
\begin{minipage}[t]{0.466\textwidth}
  \begin{center}
  Sascha Metz\\
  \normalfont{Technische Universität Darmstadt}\\
  \texttt{smetz1405@gmail.com}
  \end{center}
\end{minipage}
  \And
\begin{minipage}[t]{0.466\textwidth}
  \begin{center}
  Sarah Boufelja\\
  \normalfont{IBM}\\
  \texttt{boufelja.sarah@gmail.com}
  \end{center}
\end{minipage}
  \And
\begin{minipage}[t]{0.466\textwidth}
  \begin{center}
  Sylvester Kaczmarek\\
  \normalfont{Imperial College London}\\
  \texttt{sylvester.kaczmarek@gmail.com}
  \end{center}
\end{minipage}
  \And
\begin{minipage}[t]{0.466\textwidth}
  \begin{center}
  Klaus Merz\\
  \normalfont{European Space Agency}\\
  \texttt{klaus.merz@esa.int}
  \end{center}
\end{minipage}
  \And
\begin{minipage}[t]{0.466\textwidth}
  \begin{center}
  José A.~Martinez-Heras\\
  \normalfont{European Space Agency}\\
  \texttt{jose.antonio.martinez.heras@esa.int}
  \end{center}
\end{minipage}
  \And   
\begin{minipage}[t]{0.466\textwidth}
  \begin{center}
  Francesca Letizia\\
  \normalfont{European Space Agency}\\
  \texttt{francesca.letizia@esa.int}
  \end{center}
\end{minipage}
  \And
\begin{minipage}[t]{0.466\textwidth}
  \begin{center}
  Christopher Bridges\\
  \normalfont{University of Surrey}\\
  \texttt{c.p.bridges@surrey.ac.uk}
  \end{center}
\end{minipage}
  \And 
\begin{minipage}[t]{0.466\textwidth}
  \begin{center}
  Atılım Güneş Baydin\\
  \normalfont{University of Oxford}\\
  \texttt{gunes@robots.ox.ac.uk}
  \end{center}
\end{minipage}
}

\begin{document}

\maketitle

\begin{abstract}
After decades of space travel, low Earth orbit is a junkyard of discarded rocket bodies, dead satellites, and millions of pieces of debris from collisions and explosions. Objects in high enough altitudes do not re-enter and burn up in the atmosphere, but stay in orbit around Earth for a long time. With a speed of 28,000 km/h, collisions in these orbits can generate fragments and potentially trigger a cascade of more collisions known as the Kessler syndrome. This could pose a planetary challenge, because the phenomenon could escalate to the point of hindering future space operations and damaging satellite infrastructure critical for space and Earth science applications. As commercial entities place mega-constellations of satellites in orbit, the burden on operators conducting collision avoidance manoeuvres will increase. For this reason, development of automated tools that predict potential collision events (conjunctions) is critical. We introduce a Bayesian deep learning approach to this problem, and develop recurrent neural network architectures (LSTMs) that work with time series of conjunction data messages (CDMs), a standard data format used by the space community. We show that our method can be used to model all CDM features simultaneously, including the time of arrival of future CDMs, providing predictions of conjunction event evolution with associated uncertainties.
\end{abstract}

\section{Introduction}

The risk of collisions between man-made objects in space is growing. Moreover, due to the increased amount of space debris, the expected growth of the space sector, and the planned launch of megaconstellations, the situation is expected to worsen \cite{rossi1997long}, \cite{kessler1991collisional}, \cite{anselmo1999updated}, \cite{le2018space}, \cite{virgili2016risk}. With the growth of space debris population, a few collisions between objects in orbit might trigger a chain reaction that could pollute the orbit to the point of making it inaccessible, a scenario known as the Kessler syndrome \cite{kessler1978collision}, \cite{liou2008instability}. Unless appropriate techniques are developed to handle this problem, this could severely endanger the satellite infrastructure that is increasingly essential in Earth sciences applications. Indeed, besides being fundamental to space sciences, navigation, and telecommunication, satellite constellations are nowadays central in a wide range of geoscience-related applications in climate science, Earth observation, disaster detection, and monitoring of oceans, ice, groundwater, and landmasses \cite{board2019thriving}.

The current space population in Earth's orbit is monitored via the global Space Surveillance Network (SSN): a network of optical and radar sensors handled by the US Strategic Command (USSTRATCOM). This network observes and tracks man-made objects in space, while producing a publicly available running catalog of unclassified space objects.\footnote{\url{https://www.space-track.org/} (September 2020).} For collision assessment purposes, SSN observations are fed to a propagator (i.e., a physics simulator that predicts the evolution of the state of the objects over a time period, typically seven days) and each satellite (usually called \emph{target}) is screened against the rest of the catalogue population in order to detect close approaches, which are referred to as ``conjunctions''. If a conjunction is predicted between the target and another object (usually called \emph{chaser}), a conjunction data message (CDM) is automatically issued to the owner/operator (O/O) of the satellite, containing information about the event at the time of closest approach (TCA), such as the probability of collision and covariances of the state variables of the two objects. Furthermore, information relevant to how all the above data are determined can be also present (e.g., details of the orbit simulator and dynamical environment used). For the week leading up to TCA, more SNN observations are collected and screened and CDM updates are issued with updated information about TCA (usually around three CDMs per day) to the O/O as the conjunction event evolves. O/Os are generally alerted to the event and, up to one day prior to TCA, they must make a decision about whether to perform collision-avoidance manoeuvres, which are expensive due to resource constraints. The last CDM received is usually considered the best knowledge of the state of target and chaser at TCA. The collision risk assessment during this decision process is currently performed manually. Several methods exist to establish if pairs of objects lead to a conjunction event in the propagated time-interval \cite{hoots1984analytic}, \cite{klinkrad2010space} and to obtain collision risk estimates \cite{merz2017current, braun2016operational}.  

A major barrier to the application of machine learning techniques to aid the collision management process has been the lack of publicly available large-scale CDM datasets due to the presence of sensitive information about the assets of private companies and governments. The only publicly available CDM dataset, which we call the Kelvins dataset, has been provided by the European Space Agency (ESA) in the context of the Spacecraft Collision Avoidance Challenge,\footnote{\url{https://kelvins.esa.int/collision-avoidance-challenge/} (August 2020)} where participants developed techniques to predict the final risk of collision for each event. The dataset contains CDMs collected by ESA from 2015 to 2019, with modifications to anonymise the data (e.g., absolute time information and some of the orbital state elements are not given---only relative states). 

In this work we develop a recurrent neural-network architecture that works with sequences of CDMs representing conjunction events, where the network operates on the whole set of numerical features present in the CDM format. Specifically, our architecture is a stacked LSTM\footnote{Long short-term memory} \cite{hochreiter1997long} trained in a Bayesian deep learning \cite{mackay1992practical,WhatUncertainties,MCDropout,wilson2020case} setting in order to provide uncertainty information at test time. We train this network using the Kelvins dataset. This model can be used in two modes: (1) predicting the contents of the next CDM given all previous CDMs in an event currently unfolding; (2) predicting the whole sequence of future CDMs until TCA, by feeding each predicted CDM to back to the network and predicting the next, until the last CDM is predicted by the network. These modes are analogous to the usage of probabilistic language models, where complete sentences can be sampled from a generative model primed with some initial symbols (characters or words) \cite{bengio2003neural,mnih2012fast}.

\section{Related Work}
\paragraph{Conjunction Management}

Starting from the CDMs, current collision avoidance strategies include a collision risk assessment between pairs of objects, optionally augmenting the data with external information sources. Typical risk assessment procedures have two steps. First, the states of the two objects at the time of closest approach are predicted. This prediction is generally represented by means of a probability distribution over the state variables. Then, the information contained in such distributions is leveraged to evaluate the probability of collision. Sometimes, CDMs might contain information about probability of collision and methods used for computing it.

In the current practices, propagation is performed by either using linear/linearised methods (e.g., Kalman filter) \cite{vallado2001fundamentals}, or by using simulation-based techniques (e.g., Monte Carlo simulation) \cite{ford2005quantifying}. Semi-analytic techniques have been also developed in order to trade-off the accuracy of simulation-based techniques with the speed of linearised methods. In this context, some of the most popular semi-analytic techniques are: differential algebra with Gaussian mixture model \cite{sun2019nonlinear}, polynomial chaos expansion \cite{jones2013nonlinear}, generalized polynomial algebra \cite{vasile2019set}, unscented transform \cite{adurthi2015conjugate}.

Given the output of the propagation, the collision risk (i.e., the probability that the two objects will collide at time of closest approach) can be evaluated using several different algorithms \cite{alfriend1999probability, patera2001general, klinkrad2010space,patera2003satellite,chan2008spacecraft}. The operators then use this risk information combined to their own risk assessment and evaluation to decide whether to maneuver the satellite for avoiding conjunctions.

\paragraph{Bayesian Deep Learning and Monte Carlo Dropout}
The problem of quantifying the uncertainty associated with the estimates produced by machine learning methods is paramount for their usage in safety-critical applications \cite{klloss, LaserNetAEUncertainty, WhatUncertainties} like satellite conjunction management. Typical deep learning approaches produce only point estimates, without providing a quantitative assessment of the reliability of the prediction being made. Given the lack of interpretability of deep learning approaches, it is difficult to ascertain whether the space operator should trust the outputs of the neural network. For this reason, Bayesian deep learning methods output distributions of predictions, which allow to quantify this uncertainty (e.g., in terms of entropy or variances of the output \cite{WhatUncertainties}). This is attained by placing a prior distribution $p(\omega)$ over the parameters $\omega$ of a differentiable function implemented with a neural network $f$, and by finding the posterior distribution $p({\omega}|X,Y)$ (where $X,Y$ represent the training dataset input and output respectively) that capture the most likely functions given data, which can be used to predict the output $y^*$ for a new input $x^*$. 
The predictive distribution is: $p(y^*|x^*,X,Y) = \int p(y^*,f^*) p(f^*|x^*,\omega)p(\omega|X,Y)df^* d\omega$. Unfortunately, $p(\omega|X,Y)$ cannot be evaluated analytically. To tackle this issue, variational inference is applied: an easy-to-evaluate variational distribution $q(\omega)$ is defined, and used to approximate $p(\omega|X,Y)$ by minimizing the KL divergence $KL(q(\omega)||p(\omega|X,Y))$ resulting in the approximate predictive distribution: $ p(y^*|x^*,X,Y) = \int p(y^*,f^*) p(f^*|x^*,\omega)q(\omega)df^* d\omega$.

Many variational approximation schemes are possible. In this paper, we choose an approach called Monte Carlo dropout \cite{MCDropout, Gal2015BayesianCN} which has been popular in many fields (e.g. autonomous driving \cite{WhatUncertainties}, medical diagnosis \cite{MCDropoutDiagnosis}) thanks to its implementation simplicity. In this setting applying dropout at training time is equivalent to performing variational inference using a Bernoulli approximating distribution. In order to obtain distributions over new data points $y^*$, it is sufficient to run the network multiple times on the same input with the dropout at test time.

\section{Conjunction Event Evolution Analysis with Bayesian Deep Learning}

\begin{figure} 
    \centering
    \includegraphics[width=\textwidth]{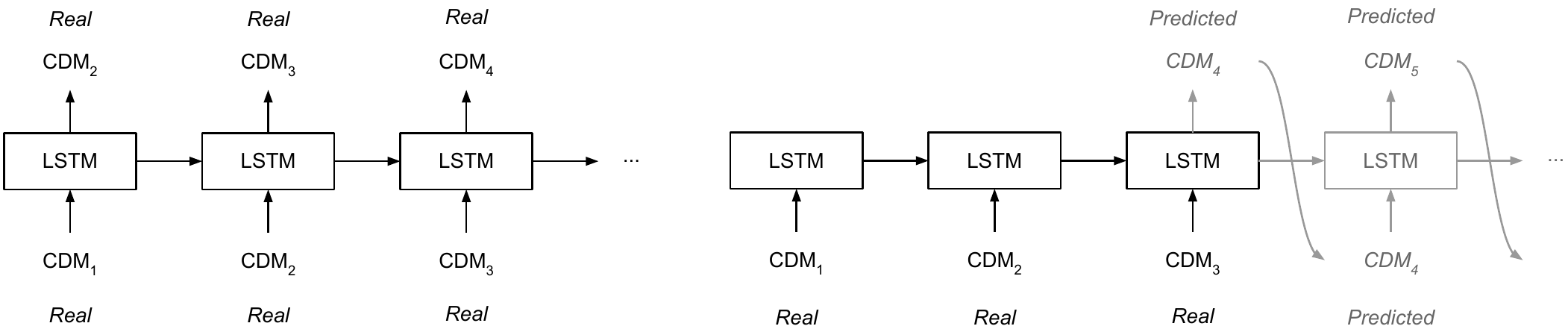}
    \caption{\emph{Left:} At training time the LSTM is optimized to reproduce the time-shifted CDM sequences in training data. \emph{Right:} At test time, given a sequence of CDMs for the current event, the trained LSTM is used to predict future CDMs that characterise the event evolution.}
    \label{fig:lstm_train_test}
    \vspace{-4mm}
\end{figure}

While the contestants of the Spacecraft Collision Avoidance Challenge were required to predict the final risk and perform binary classification (i.e., high- versus low-risk event) \cite{Kelvins}, this approach does not reflect the two-step nature of the established risk assessment procedures. In particular, the risk prediction and the decision of whether to classify a CDM sequence as either high or low risk is made by a neural network. This decision cannot be explained in terms of physical variables, and hence cannot be trusted by operators. Indeed, the probability of collision is a quantity obtained by integrating physical quantities contained in CDMs (e.g., using \cite{alfriend1999probability, patera2001general, klinkrad2010space,patera2003satellite,chan2008spacecraft}) and its evaluation is carefully designed to capture the uncertainty contained in the distributions of the states of the two objects at TCA. Another problem with a classification approach in this setting is the severe data imbalance \cite{wang2012imbalance} between the high- and low-risk classes, since high-risk events and collisions are very rare in actual real data. For these reasons, we propose to use generative modeling to model distributions of future CDMs conditioned on previous CDMs in an event, and include all the numerical features within the CDM data format in model inputs and outputs. This allows us to fully make use of the available training data to aid representation learning \cite{bengio2013representation}, and also grounds our approach in the already established CDM data format for which there are tools and techniques developed by the space community. In particular, we employ a stack of LSTMs trained by minimising the mean squared error (MSE) between predicted next CDMs and the ground-truth, using each event's CDM sequence as input and the same sequence shifted by one time step as the target, so that in each time step the network learns to predict the next CDM conditioned on all previous CDMs (Figure~\ref{fig:lstm_train_test}, left). 

In order to produce a distribution over predicted CDMs, representing uncertainty, we apply Monte Carlo dropout \cite{MCDropout} and sample multiple predictions at test time. Referring to a conjunction event (a CDM sequence) as $\gamma_{1:t}=\{\gamma_1,\dots,\gamma_t\}$, where $\gamma_t$ is the CDM at time $t$, the network at test time allows us to sample the next predicted CDM $\bar{\gamma}_{t+1} \sim p(\bar{\gamma}_{t+1}\vert\gamma_{1:t})$ conditioned on $\gamma_{1:t}$, the CDMs since the beginning of the event up to current time $t$. In addition, the predictions can be extended further into the future by sampling subsequent $\bar{\gamma}_{t+2} \sim p(\bar{\gamma}_{t+2}\vert\gamma_{1:t},\bar{\gamma}_{t+1})$, and then $\bar{\gamma}_{t+3} \sim p(\bar{\gamma}_{t+3}\vert\gamma_{1:t},\bar{\gamma}_{(t+1):(t+2)})$ and so on, until a complete event prediction until TCA is obtained as $\bar{\gamma}_{1:T} = \{\gamma_1,\dots,\gamma_t,\bar{\gamma}_{t+1},\dots,\bar{\gamma}_{T}\}$ (Figure~\ref{fig:lstm_train_test}, right).

\section{Experiments}

For our experiments, we use the Kelvins dataset,\footnote{Description of data features: \url{https://kelvins.esa.int/collision-avoidance-challenge/data/}} and discard the features \verb+c_rcs_estimate+, \verb+t_rcs_estimate+ (as these contain many NaN values and we used \verb+c_span+, \verb+t_span+ to leverage the size of the objects), the parameters related to the solar flux (e.g., \verb+F10+, \verb+F3M+, \verb+SSN+, \verb+AP+) because they also contain many NaN values and are not normally included in CDMs (i.e., they have been augmented by the organizers of the competition). Furthermore, we drop \verb+mission_id+, \verb+c_object_type+ because they remain constant through the whole event sequence; moreover, the \verb+mission_id+ cannot generalize when applied to test data with other mission ids. Then we drop all the CDMs with NaN values, and CDMs with abnormal values in the variances of the state variables (i.e., \verb+t_sigma_r+ $ > 20$, \verb+c_sigma_r+ $ > 1000$, \verb+t_sigma_t+ $ > 2000$, \verb+c_sigma_t+ $ > 100000$, \verb+t_sigma_n+ $ > 10$, \verb+c_sigma_n+ $ > 450$). After these steps, given the wide scale variation among the variables involved, to stabilise the training, we normalize each feature (except the categorical ones) by subtracting the mean and dividing the result by the standard deviation. For the next CDM prediction approach, we consider only event sequences with at least two CDMs. Starting from the 199,082 CDMs in the original dataset, after the cleaning procedure 156,668 CDMs remain, which are grouped into 11,387 events. We randomly split the full Kelvins dataset into a training and test split with the test split amounting for the 15\% of the total data (9,367 events in the training set, with 133,435 CDMs). With this data, we train a network that handles 52 CDM features in total, including miss distance, relative velocity and position, and target and chaser covariance matrices.\footnote{Note that when training with CDM datasets other than Kelvins, there can be more features which were not available in Kelvins data. We are planning to show examples of this in an upcoming paper.} The model used is a stack of two 256-dimensional LSTMs, whose hidden state is passed through a ReLU activation and fed to a linear layer to obtain the final prediction. Dropout is applied to all but the output layers of the network with 0.2 dropout rate. The overall model has 857,140 parameters, and is trained for 500 epochs with the Adam optimizer \cite{adam}, learning rate $10^{-4}$ and batch size 128. We will make the code of our implementation publicly available.

\begin{figure}
    \centering
    \includegraphics[width=\textwidth]{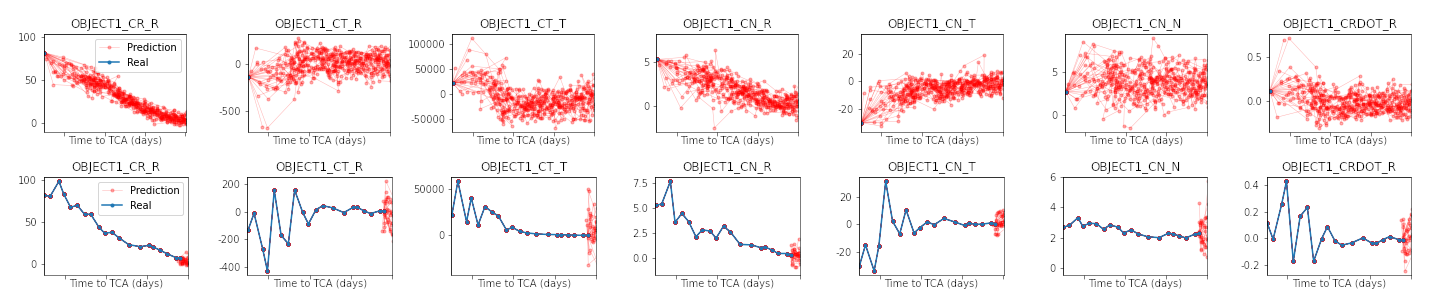}
    \caption{\emph{Top row:} the evolution of the values of a selection of CDM features through a whole event; the prediction is made using only the first CDM. \emph{Bottom row:} as new CDMs arrive, the predictions made using only he first CDM are confirmed (blue dots), showing that the LSTM can predict non-trivial patterns describing event evolution.}
    \label{fig:uncertainty_evolution}
\end{figure}

We report the mean squared error (MSE) of a baseline that predicts the variables at the next time step to be the same of the previous time step. While using the risk contained in the last CDM received was the core strategy leveraged by the participants to the Space Collision Avoidance challenge \cite{Kelvins} to discriminate between high and low risk events, in order to predict the physical quantities involved, it performs poorly ($MSE = 0.2433$). Even only with one Monte Carlo dropout sample we obtain better performance ($MSE = 0.1753$). With $n = 50$ samples, we obtain further improvements ($MSE = 0.1419$). An important remark is that, since the arrival time of the next CDM is modelled as one of the variables predicted, our model can give a prediction also in terms of when the next CDM will arrive, further informing the operators about the decision to be made. Figure \ref{fig:uncertainty_evolution} gives an example event evolution from the validation set, showing that the model has learned non-trivial patterns about the behaviour of the physical variables involved.

\section{Conclusion}
In this work we introduced a new recurrent neural network architecture to model the time evolution of conjunction events, using sequences of the standard CDM data format in use by the space community. We demonstrated that non-trivial time evolution dynamics of conjunction events can be learned, and the trained models can be used to predict event outcomes at test time. Furthermore we applied Monte Carlo dropout in order to produce distributions over the CDM features describing uncertainties associated with the two monitored objects at the time of closest approach. This method opens up a new way of working with CDM data sequences using machine learning, which we believe is an essential step in the automation and scaling of conjunction management operations in the near future.

\section{Broader Impact}
Given the increasing amount of space debris and the plans for new mega-constellations of satellites in orbit, the problem of automated collision management will become paramount to preserve the global satellite infrastructure network that currently provides critical services (e.g., weather data monitoring, global communications, Earth observation). Furthermore, collisions in the increasingly crowded low Earth orbit might trigger a cascade of collisions (Kessler Syndrome) that could hinder future space operations and trap us on Earth for years to come. Hence it is crucial to develop tools that will help maintain a managed low Earth orbit population. We do not foresee our research to have negative ethical impacts on society.

\section*{Acknowledgments}
This work has been enabled by Frontier Development Lab (FDL) Europe, a public--private partnership between the European Space Agency (ESA), Trillium Technologies and the University of Oxford, and supported by Google Cloud. We would like to thank Dario Izzo and Moriba Jah for sharing their technical expertise and James Parr, Jodie Hughes, Leo Silverberg, Alessandro Donati for their support. AGB is supported by EPSRC/MURI grant EP/N019474/1 and by Lawrence Berkeley National Lab.

\bibliographystyle{unsrt}
\bibliography{main.bib}

\end{document}